\documentclass{article}

\PassOptionsToPackage{numbers,sort&compress}{natbib}
\usepackage[preprint]{neurips_2026}

\usepackage[utf8]{inputenc}
\usepackage[T1]{fontenc}
\usepackage{courier} 
\usepackage{hyperref}
\usepackage{url}
\usepackage{booktabs}
\usepackage{amsmath}
\usepackage{amsfonts}
\usepackage{graphicx}
\usepackage{microtype}
\usepackage{xcolor}
\usepackage{tikz}
\usetikzlibrary{arrows.meta,positioning,fit}

\definecolor{reportblue}{RGB}{62,117,176}
\definecolor{measureorange}{RGB}{230,126,34}
\definecolor{fusiongreen}{RGB}{64,145,108}
\definecolor{lightgray}{RGB}{245,245,245}

\title{Integrating Heterogeneous Echocardiography AI Outputs in Language Space:
Learnable Late Fusion with a Supervised LLM}

\author{%
  \textbf{Ryo Takizawa \qquad Tempei Kabayama \qquad Chiemi Yamazaki \qquad Shun Kitamura} \\[0.4em]
  \textbf{Ryo Matsuoka \qquad Yuta Ando \qquad Satoshi Kodera \qquad Norihiko Takeda} \\[0.6em]
  The University of Tokyo Hospital \\
  \href{mailto:takizawa@isi.imi.i.u-tokyo.ac.jp}{\texttt{takizawa@isi.imi.i.u-tokyo.ac.jp}}%
}

\begin{document}

\maketitle

\begin{abstract}
View- and target-specific echocardiography AI models capture useful but complementary information
and may not achieve sufficient predictive performance in isolation. A learned ensemble is a natural
way to combine these outputs, but heterogeneous models typically require each output to be
converted into a binary decision, probability, confidence score, or fixed-length feature vector. We
developed a supervised large language model (LLM)-based late-fusion module that directly aggregates
view-specific predicted finding statements and predicted measurements in language space and learns
a case-level fusion function. In patient-level three-fold cross-validation of 859 examinations from
646 patients, the proposed method integrated heterogeneous outputs within a single framework,
performed comparably to the best-performing reference strategy evaluated, and outperformed a
general-purpose LLM without task-specific training. These results suggest that a supervised LLM can
combine multiple upstream models without model-specific output conversion or feature engineering
that requires medical knowledge.
\end{abstract}

\section{Introduction}

Multiple echocardiographic views provide complementary information about cardiac morphology and
function. Deep-learning-based echocardiographic analysis has progressed from view classification
and chamber segmentation to ejection-fraction estimation and disease or finding classification
\citep{zhang2018fully,ouyang2020video}. More recently, vision--language foundation models trained
on large collections of paired videos and reports, as well as methods that address many
interpretation tasks within a single model, have been reported
\citep{christensen2024vision,holste2025complete}. Nevertheless, individual models currently may not
provide adequate examination-level performance, motivating the integration of complementary
evidence captured by different models.

A conventional learned ensemble is stacked generalization, in which a meta-learner is trained on
the outputs of upstream models \citep{wolpert1992stacked}. Meta-learners such as logistic
regression, gradient boosting, and multilayer perceptrons can learn the reliability of multiple
models and correlations among their errors. Connecting heterogeneous upstream models to a common
meta-learner, however, generally requires defining each output as a binary decision, probability,
confidence score, or fixed-length numerical feature. When an upstream model or evaluation target
changes, the corresponding output layer, threshold, feature schema, and missing-value handling must
often be redesigned.

Echocardiography AI outputs may contain a mixture of text describing severity and uncertainty,
continuous measurements, units, and missing items. Binarizing these outputs or converting them into
fixed-length features in advance can discard information. An LLM can place the original outputs of
different models into a common context comprising natural language and numbers---that is, language
space---and learn a fusion function without model-specific binarization or feature engineering. This
common interface may reduce feature engineering when upstream models or input items are added or
changed. LLMs have begun to be used as orchestrators that combine the outputs of multiple medical
AI tools \citep{heidari2025echocopilot,taratynova2026echosonar}, but it remains unclear whether
task-specific supervised learning can adapt the fusion process to the error distributions of
particular upstream models.

We developed a supervised MedGemma-based LLM fusion module that jointly classifies four findings at
the examination level: left ventricular (LV) dilatation, left atrial (LA) dilatation, aortic root
dilatation, and global LV systolic dysfunction. Upstream report models predict finding statements,
including presence and severity (e.g., \texttt{mild}, \texttt{moderate}, and \texttt{severe}), from
as many as five echocardiographic views. Measurement models predict LV end-diastolic diameter
($D_d$), LA diameter (LAD), sinus of Valsalva diameter, biplane Simpson ejection fraction
(EF$_{\mathrm{MOD}}$), and Teichholz ejection fraction (EF$_{\mathrm{Teichholz}}$) from the
corresponding views. The proposed
method directly places these predicted statements and measurements into a shared text context. We
investigated whether a single learnable framework could integrate heterogeneous upstream outputs
without manually mapping each output to a target or designing conversions to binary decisions or
confidence scores. Patient-level cross-validation evaluated predictive performance, behavior in the
presence of weak inputs, and performance relative to a general-purpose LLM unable to learn a fusion
strategy tailored to the upstream models.

Our main contributions are threefold:
\begin{enumerate}
    \item We construct a supervised LLM late-fusion module that directly integrates view-specific
      predicted finding statements and predicted measurements in language space, without
      model-specific engineering.
    \item We evaluate simultaneous prediction of four targets from multiple imperfect upstream
      models using patient-level three-fold cross-validation of 859 examinations from 646 patients.
    \item We characterize the value and limitations of a supervised LLM as a fusion module in terms
      of a common representation for heterogeneous outputs, behavior with weak inputs, and the
      potential to extend the input configuration.
\end{enumerate}

\section{Related work}

\subsection{Automated echocardiography analysis and model integration}

Echocardiography AI has expanded from individual perception tasks toward examination-level
interpretation. \citet{zhang2018fully} reported an automated pipeline spanning view identification,
segmentation, quantification, and disease detection across multiple standard views. EchoNet-Dynamic
performs LV segmentation, ejection-fraction estimation, and cardiomyopathy classification from
echocardiography videos, demonstrating the benefit of temporal information from the entire video
\citep{ouyang2020video}. EchoCLIP demonstrated the utility of vision--language pretraining on a
large collection of paired echocardiographic video frames and reports \citep{christensen2024vision}, while
PanEcho addressed numerous interpretation tasks across diverse views within a single model
\citep{holste2025complete}. These studies focus on representation learning in an upstream model or
end-to-end examination interpretation. In contrast, we study learned downstream fusion of several
individually imperfect upstream models with heterogeneous output formats.

\subsection{Learned ensembles and fusion in language space}

Stacking is a general framework in which upstream-model predictions are inputs to a meta-learner
that
learns a fusion function \citep{wolpert1992stacked}. Although it can combine models to outperform
individual predictors, conventional meta-learners require upstream outputs to be represented as
consistent numerical features. For a classification model, one might select a binary prediction or
probability; for a regression model, a continuous value; and for a generative model, an extracted
finding or embedding. A fixed-length schema must then be designed for the model configuration. This
preprocessing adds design work when heterogeneous models are connected or updated and may discard
severity, negation, or ambiguity expressed in text.

The central motivation for using an LLM in this work is its ability to directly process finding
statements, numbers, units, missing items, and a variable number of outputs in a common
natural-language input space. The upstream models need not expose standardized probabilities or
confidence scores, and their original predictive representations are preserved while
examination-level fusion is learned. Echo-CoPilot uses multiple LLM-based agents to invoke
specialized echocardiography tools and reconcile structural, pathological, and quantitative
perspectives \citep{heidari2025echocopilot}, while EchoSonar-R unifies classification and report
generation across multiple views \citep{taratynova2026echosonar}. Our work differs by neither
providing images directly to the LLM nor allowing it to select tools at inference time. Instead, it
applies supervised late fusion to fixed outputs from existing view-specific report and measurement
models.

\section{Method}

\subsection{Problem setting and dataset}

This retrospective observational study used echocardiography examinations acquired at a single
center. The analysis included 859 examinations from 646 patients obtained between January 2015 and
December 2023. Of the 646 patients, 135 had more than one examination. Additional details are provided in Appendix~\ref{app:cohort}.

\subsection{Reference labels}

Reference labels were extracted from predefined findings in clinical reports written by physicians
during routine care. The four evaluation targets were (1) LV dilatation, (2) LA dilatation, (3)
aortic root dilatation, and (4) global LV systolic dysfunction, each represented as positive or
negative.

Mild-or-greater enlargement was considered positive for the LV and LA, and aortic root dilatation
was considered positive when explicitly documented. Global LV systolic dysfunction was considered
positive on the basis of prespecified descriptions such as diffuse LV wall-motion reduction.
Descriptions corresponding to \emph{normal}, \emph{absent}, \emph{trace}, \emph{borderline}, or
\emph{low-normal} were classified as negative.

\subsection{Upstream model outputs}

\subsubsection{View-specific predicted finding statements}

The upstream report prediction model was a multitask classifier that predicted the severity or type
of predefined findings separately from apical four-chamber (4CH), apical two-chamber (2CH), apical
three-chamber (3CH), parasternal long-axis (LAX), and parasternal short-axis (SAX) views.
Anatomically informative views differed by target. Predictions were available from 4CH, 2CH, and
LAX for LV and LA dilatation; from 3CH and LAX for aortic root dilatation; and from 4CH, 2CH, 3CH,
and SAX for global LV systolic dysfunction.

The report model's VideoEncoder used a ViViT-type encoder pretrained by the Video CLIP method of
\citet{takizawa2025videoclip}. During pretraining, the encoder received a fixed-length sequence of 32 frames from
each video and produced a 512-dimensional video embedding. A BERTJapaneseV3 text encoder processed
Japanese clinical reports of up to 256 tokens. The two encoders were trained with a bidirectional
contrastive loss that treated matched video--report pairs as positives. Pretraining used 747,900
videos and examination-level clinical reports from 60,747 examinations of 29,886 patients across
the five views \citep{takizawa2025videoclip}.

For the upstream report model used here, view-specific multitask classification heads were attached
to the pretrained VideoEncoder to predict multiple findings observable in a single echocardiography
video. The highest-probability class for each finding was selected and deterministically converted
to an English finding statement using a predefined mapping. For late fusion, statements were
concatenated
by view; class probabilities were not
used. Thus, report inputs to the LLM were templated statements that retained the type and severity
of a finding, including terms such as \texttt{mild}, \texttt{moderate}, and \texttt{severe}, rather
than freely generated text.

Report predictions were generated in advance and held fixed during late-fusion training and
evaluation; the upstream report model parameters were not updated. The late-fusion cohort excluded
all examinations and patients used to train the report model.

\begin{figure}[t]
  \centering
  \includegraphics[width=\linewidth]{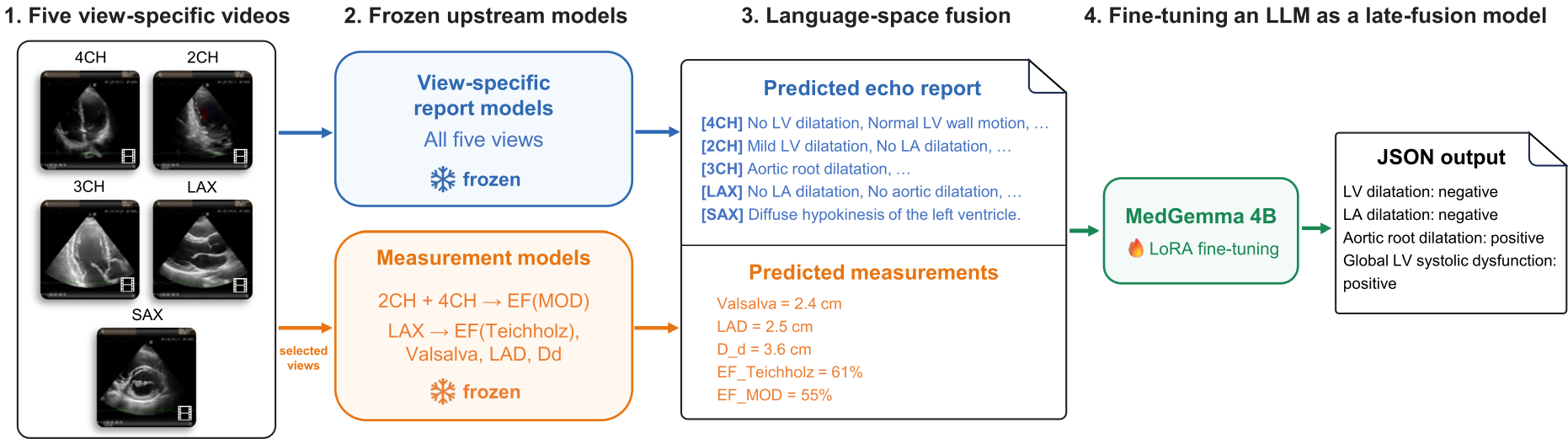}
  \caption{Study pipeline. View-specific report models produce finding statements, and measurement
    models produce continuous values. These heterogeneous outputs are formatted as
    examination-level text and passed to a supervised MedGemma-based fusion module, which jointly
    returns four binary findings in a fixed JSON schema.}
  \label{fig:pipeline}
\end{figure}

\subsubsection{Predicted measurements}

The measurement model combined semantic segmentation of 4CH, 2CH, and LAX videos with geometric
postprocessing that involved no additional learning. For each view, a segmentation
model consisting of DeepLabV3 \citep{chen2017rethinking} with an ImageNet-pretrained ResNet-50 backbone and a
lightweight U-Net-style upsampling module was transfer learned independently. The 4CH model
delineated the myocardial and blood-pool region (MASS), LV, LA, right ventricle (RV), and right atrium (RA);
the 2CH model delineated MASS, LV, and LA; and the LAX model delineated the LV, interventricular septum (IVS),
LV posterior wall (LVPW), aorta (AO), and LA.

Frame-wise segmentation logits were first smoothed temporally with a Gaussian moving average.
End-diastolic (ED) and end-systolic (ES) candidates were detected from the time series of LV and LA
areas: local maxima of LV area were ED candidates, and local minima were ES candidates. The minimum
inter-peak interval was 0.5~s and the ED--ES search window was 0.1--0.5~s. An ES candidate was
additionally required to have an LA area above the cardiac-cycle mean within a $\pm0.1$-s window.
When multiple ED--ES pairs were available, the pair whose derived EF was closest to the reference
value of 62.5\% was selected.

For ED and ES frames in 4CH and 2CH, contours were extracted from the LV probability map with
marching squares and smoothed using a low-pass Fourier-descriptor filter. When the LV contour was
inadequate, the procedure fell back to the contour of the MASS region containing the LV. The LV
inflow corresponding to the mitral annulus was estimated from proximity to the LA contour, and the
most distant point on the LV contour was designated the apex. The line joining these points defined
the LV long axis. After rotating the contour to make the long axis horizontal, we divided the axis
into 20 equal segments. The upper and lower contour portions were fitted separately with
polynomials at each position, and the distance between the curves defined the disk diameter. Pixel
lengths were converted to centimeters using pixel spacing in the DICOM metadata.

For the biplane method of disks (MOD) using 2CH and 4CH, the long-axis lengths and 20 disk
diameters obtained within an examination were averaged separately for ED and ES. We used the larger
of the mean long-axis lengths from the two views in diastole and the smaller in systole:
\begin{equation}
L_d=\max(\bar L_{d,\mathrm{4CH}},\bar L_{d,\mathrm{2CH}}),\qquad
L_s=\min(\bar L_{s,\mathrm{4CH}},\bar L_{s,\mathrm{2CH}}).
\end{equation}
Treating the product of the corresponding diameters in the two views as an elliptical
cross-section, we computed
\begin{align}
\mathrm{EDV}_{\mathrm{MOD}}
&=\frac{\pi}{4}\frac{L_d}{20}\sum_{i=1}^{20}
  \bar d_{d,\mathrm{4CH}}^{\,i}\bar d_{d,\mathrm{2CH}}^{\,i}, \\
\mathrm{ESV}_{\mathrm{MOD}}
&=\frac{\pi}{4}\frac{L_s}{20}\sum_{i=1}^{20}
  \bar d_{s,\mathrm{4CH}}^{\,i}\bar d_{s,\mathrm{2CH}}^{\,i},
\end{align}
where $\bar d_{d,v}^{\,i}$ and $\bar d_{s,v}^{\,i}$ are the mean within-examination diameters of
disk $i$ in view $v$. Finally,
$\mathrm{EF}_{\mathrm{MOD}}
=(\mathrm{EDV}_{\mathrm{MOD}}-\mathrm{ESV}_{\mathrm{MOD}})
/\mathrm{EDV}_{\mathrm{MOD}}\times100$.

In LAX, the LV, IVS, LVPW, AO, and LA were treated as band-like
structures bounded by two opposing contours. Candidate shape corners were extracted at local maxima
of contour curvature. Boundaries were divided according to segment length and the area of the
enclosing polygon, after which their anatomical orientations were aligned using the centroids of
adjacent structures. Dynamic time warping established point correspondences between opposing
boundaries, and the distances between matched points formed a local diameter profile. The LV
internal diameter was measured 2~cm apical to the position corresponding to the mitral leaflet
tips, yielding $D_d$ at ED and $D_s$ at ES. The maximum diameter of the profile corresponding to
the aortic sinus defined the sinus of Valsalva diameter, and the diameter near the LA contour
centroid defined LAD.

Teichholz volumes were calculated from $D_d$ and $D_s$ in centimeters:
\begin{equation}
\mathrm{EDV}_{\mathrm{Teichholz}}=\frac{7}{2.4+D_d}D_d^3,\qquad
\mathrm{ESV}_{\mathrm{Teichholz}}=\frac{7}{2.4+D_s}D_s^3,
\end{equation}
and
$\mathrm{EF}_{\mathrm{Teichholz}}
=(\mathrm{EDV}_{\mathrm{Teichholz}}-\mathrm{ESV}_{\mathrm{Teichholz}})
/\mathrm{EDV}_{\mathrm{Teichholz}}\times100$.
For late-fusion inputs, the values were listed as text in the order of EF$_{\mathrm{Teichholz}}$, Valsalva sinus diameter,
LAD, $D_d$, and EF$_{\mathrm{MOD}}$ for each examination.

The measurements associated with the four targets were $D_d$ for LV dilatation, LAD for LA
dilatation, sinus of Valsalva diameter for aortic root dilatation, and EF$_{\mathrm{MOD}}$ and
EF$_{\mathrm{Teichholz}}$ for global LV systolic dysfunction (see
Appendix~\ref{app:signals} for details).
Measurements were computed and fixed
before late-fusion training, and the measurement-model parameters were not updated. The late-fusion
cohort excluded all examinations and patients used to train the measurement model.

\subsection{Fusion task in language space}

For each examination, all available view-specific predicted finding statements and measurements
were converted to templated text that distinguished the view, measurement name, value, and unit
(Figure~\ref{fig:pipeline}).
Report predictions remained as finding statements, and measurement predictions remained continuous.
We neither binarized upstream outputs in advance nor required upstream models to provide
probabilities or confidence scores. A system prompt supplied the LLM with the positive and negative
definitions for each target, and the model returned all four targets simultaneously in a fixed JSON
schema. The output was parsed into four separate binary decisions. The exact system prompt,
user-prompt template, and JSON schema are provided in Appendix~\ref{app:prompts}.

\subsection{Supervised fine-tuning}

We used \texttt{google/medgemma-4b-it} \citep{sellergren2025medgemma} as the base model and applied
low-rank adaptation (LoRA) \citep{hu2022lora} for parameter-efficient fine-tuning. The target
modules were the query, key, value, and output projections in the attention layers and the gate,
up, and down projections in the feed-forward layers. The fine-tuning configuration is provided in
Appendix~\ref{app:hyperparameters}.

For every fold, all original training examinations were retained once. We then sampled, with
replacement, an additional set twice the size of the original training set, producing three times
as many training examples as in the unaugmented set. For each of the four labels, we calculated the
item-specific positive rate required in the additional sample to bring the final training-set
positive rate close to 0.30 and iteratively updated examination weights 200 times. Normalized
weights were capped at 6.0 to avoid excessive duplication of rare label combinations. Test sets
were not modified.

\section{Results}

\subsection{Evaluation setting}

We performed three-fold cross-validation with patients as the splitting unit (Table~\ref{tab:splits}). All examinations from
a patient were assigned to the same fold. The split approximately balanced the numbers of
examinations, patients, positives for each of the four targets, and observed label combinations
across folds. Every examination appeared in a test set exactly once, and predictions from the three
test folds were concatenated to form out-of-fold (OOF) predictions for the full cohort.

\begin{table}[t]
  \caption{Patient-level three-fold cross-validation splits. Positive counts refer to each test
    fold.}
  \label{tab:splits}
  \centering
  \small
  \setlength{\tabcolsep}{2.2pt}
  \begin{tabular}{rrrrrrrrr}
    \toprule
    Fold & \multicolumn{2}{c}{Train} & \multicolumn{2}{c}{Test} & LV dil. & LA dil. & Root dil. &
    LV dysf. \\
    \cmidrule(lr){2-3}\cmidrule(lr){4-5}
      & Exams & Patients & Exams & Patients & Pos. & Pos. & Pos. & Pos. \\
    \midrule
    1 & 573 & 431 & 286 & 215 & 26 & 95 & 49 & 23 \\
    2 & 572 & 430 & 287 & 216 & 26 & 95 & 49 & 23 \\
    3 & 573 & 431 & 286 & 215 & 25 & 94 & 49 & 24 \\
    \midrule
    \textbf{Overall} & -- & -- & \textbf{859} & \textbf{646} & \textbf{77} &
    \textbf{284} & \textbf{147} & \textbf{70} \\
    \bottomrule
  \end{tabular}
\end{table}


\subsection{Standalone performance of upstream models}

Table~\ref{tab:upstream} reports standalone F1 scores for each upstream signal and the
fold-specific thresholds used to binarize measurement predictions. Report predictions were
evaluated separately for each view-specific classification head. For measurements, the threshold
and decision direction selected on each outer fold's training set were fixed and applied to that
fold's test set. All F1 scores were calculated after combining the three test folds into an OOF
evaluation.

\begin{table}[t]
  \caption{Standalone performance of upstream predictions. Dashes indicate that a view did not
    output the target. Measurement thresholds maximized positive-class F1 on the corresponding
    training fold.}
  \label{tab:upstream}
  \centering
  \small
  \setlength{\tabcolsep}{2pt}
  \textbf{(a) View-specific report prediction F1}\\[2pt]
  \begin{tabular}{lccccc}
    \toprule
    Target & 4CH & 2CH & 3CH & LAX & SAX \\
    \midrule
    LV dilatation & 0.677 & 0.687 & -- & 0.687 & -- \\
    LA dilatation & 0.728 & 0.721 & -- & 0.736 & -- \\
    Aortic root dilatation & -- & -- & 0.459 & 0.547 & -- \\
    Global LV systolic dysfunction & 0.731 & 0.731 & 0.748 & -- & 0.688 \\
    \bottomrule
  \end{tabular}

  \vspace{5pt}
  \textbf{(b) Predicted measurements, fold-specific thresholds, and OOF F1}\\[2pt]
  \begin{tabular}{llcccc}
    \toprule
    Target & Measurement & Fold 1 & Fold 2 & Fold 3 & OOF F1 \\
    \midrule
    LV dilatation & $D_d$ (cm) & $\geq5.15$ & $\geq5.25$ & $\geq5.05$ & 0.634 \\
    LA dilatation & LAD (cm) & $\geq2.85$ & $\geq2.95$ & $\geq2.85$ & 0.681 \\
    Aortic root dilatation & Valsalva (cm) & $\geq2.75$ & $\geq2.75$ & $\geq2.75$ & 0.506 \\
    Global LV dysfunction & EF$_{\mathrm{MOD}}$ (\%) & $\leq42.5$ & $\leq41.5$ & $\leq42.5$ & 0.459
    \\
    Global LV dysfunction & EF$_{\mathrm{Teichholz}}$ (\%) & $\leq42.5$ & $\leq35.5$ & $\leq42.5$ &
    0.417 \\
    \bottomrule
  \end{tabular}
\end{table}

Report-prediction F1 exceeded the corresponding measurement-prediction F1 for LV dilatation, LA
dilatation, and global LV systolic dysfunction. For aortic root dilatation, performance differed
substantially between views.

\subsection{Main results}

We evaluated the proposed method from two perspectives: whether it could achieve a reasonable
fusion strategy and whether learning the fusion strategy was necessary. To this end, we defined
three classes of baseline: interpretable fixed rules, conventional learned fusion, and a
general-purpose LLM without additional training.

\paragraph{Interpretable fixed-rule baselines.}
For each target, we computed OR and AND across all report predictions from the available views,
across all measurements, and across all report and measurement signals. For example, all signals OR
was positive if any available report prediction or binarized measurement was positive, whereas all
signals AND was positive only if all were positive. Continuous measurements required by the fixed
rules were binarized using the thresholds described in Table~\ref{tab:upstream}.
Because their mappings from inputs to final decisions could be traced explicitly, these fixed rules
served as interpretable baselines for evaluating learned LLM fusion.

\paragraph{Conventional learned LightGBM baseline.}
To compare the proposed method with learned fusion corresponding to conventional stacked
generalization, we constructed a baseline using LightGBM \citep{ke2017lightgbm} as the meta-learner.
For each of the four targets, the corresponding view-specific report-prediction classes were
converted to one-hot vectors, concatenated with the corresponding continuous measurements, and
provided to an independent binary classifier. Specifically, LV dilatation used report predictions
from 4CH, 2CH, and LAX together with $D_d$; LA dilatation used report predictions from 4CH, 2CH, and
LAX together with LAD; aortic root dilatation used report predictions from 3CH and LAX together with
the Valsalva sinus diameter; and global LV systolic dysfunction used report predictions from 4CH,
2CH, 3CH, and SAX together with EF$_{\mathrm{MOD}}$ and EF$_{\mathrm{Teichholz}}$. This input design
required explicit mappings between evaluation targets and upstream outputs, as well as a
fixed-length feature schema.

The LightGBM hyperparameters were fixed in advance at 200 trees, a learning rate of 0.03, a maximum
depth of 3, a maximum of 7 leaves, a minimum of 20 child samples, and L2 regularization of 1.0. The
original outer-training examinations were not resampled, and balanced class weights were used
during training. Within the training set of each outer fold, we generated three-fold OOF predictions
using patients as the splitting unit and selected the binary decision threshold that maximized
positive-class F1. The model was then refit on the full outer-training set, and the selected threshold
was held fixed when applied to the outer-test set.

\paragraph{Strong general-purpose LLM baseline.}
Because LLMs possess medical-domain knowledge, it was not self-evident that a fusion strategy needed
to be learned. To assess the need to learn a strategy tailored to the upstream models, we provided
the same prompts used for supervised fine-tuning (SFT) to the strong general-purpose LLM \texttt{gpt-5.5} and had it classify
all examinations. Reasoning effort was set to \texttt{none}, and each examination was evaluated
once.

\paragraph{Metrics.}
The primary descriptive metric was Macro-F1, the mean positive-class F1 across the four targets.
Accuracy alone was not used as the primary metric because prevalences differed and positive
examples of LV dilatation and global LV systolic dysfunction were uncommon.
Secondary metrics were Micro-F1 pooled across all target decisions, exact match (the fraction of
examinations for which all four targets were correct), and Hamming accuracy (the fraction of all
binary decisions that were correct).
All four targets could be extracted under the fixed schema from every output of SFT all, the
report-only ablation described below, and GPT-5.5; no samples were excluded from evaluation because
of parsing failures.

Differences in Macro-F1 between methods were assessed with a paired patient-cluster bootstrap of
20,000 replicates, evaluating both methods in the same bootstrap sample. This exploratory analysis
was added after inspection of the primary results and was not a prespecified confirmatory test. We
describe whether each 95\% CI includes zero but do not interpret the analysis as a superiority or
non-inferiority test. Between-fold standard deviations are descriptive of the data subsets and
trained models and were not used for statistical inference. Because the bootstrap resampled fixed
OOF predictions, it did not capture variability from retraining models or re-estimating thresholds
under altered training data.

\begin{table}[t]
  \caption{Overall performance of the fusion methods. Bold values mark the highest value in each
    column.}
  \label{tab:overall}
  \centering
  \small
  \setlength{\tabcolsep}{2pt}
  \begin{tabular}{lccccc}
    \toprule
    Method & Macro-F1 & Micro-F1 & Exact match & Hamming acc. &
    \shortstack{Fold Macro-F1,\\mean$\pm$SD} \\
    \midrule
    Reports OR & \textbf{0.703} & \textbf{0.709} & 0.639 & 0.891 &
    \textbf{0.703$\pm$0.042} \\
    Reports AND & 0.576 & 0.587 & \textbf{0.653} & \textbf{0.894} & 0.571$\pm$0.052 \\
    Measurements OR & 0.565 & 0.586 & 0.483 & 0.832 & 0.567$\pm$0.047 \\
    Measurements AND & 0.559 & 0.598 & 0.531 & 0.851 & 0.557$\pm$0.018 \\
    All signals OR & 0.615 & 0.629 & 0.477 & 0.826 & 0.619$\pm$0.059 \\
    All signals AND & 0.473 & 0.525 & 0.637 & 0.886 & 0.469$\pm$0.048 \\
    \midrule
    LightGBM & 0.688 & 0.698 & 0.637 & 0.890 & 0.687$\pm$0.048 \\
    \textbf{LLM (SFT, all)} & 0.690 & 0.699 & 0.634 & 0.889 & 0.690$\pm$0.020 \\
    LLM (SFT, report only) & 0.693 & 0.696 & 0.633 & 0.889 & 0.693$\pm$0.033 \\
    GPT-5.5 & 0.608 & 0.631 & 0.515 & 0.848 & 0.608$\pm$0.031 \\
    \bottomrule
  \end{tabular}
\end{table}

\begin{table}[t]
  \caption{Target-specific F1. Bold values mark the highest point estimate in each row.}
  \label{tab:targetf1}
  \centering
  \small
  \setlength{\tabcolsep}{3pt}
  \begin{tabular}{lccccc}
    \toprule
    Target & Reports OR & LightGBM & SFT all & SFT report only & GPT-5.5 \\
    \midrule
    LV dilatation & \textbf{0.733} & 0.727 & 0.696 & 0.721 & 0.731 \\
    LA dilatation & 0.767 & 0.771 & \textbf{0.778} & 0.770 & 0.766 \\
    Aortic root dilatation & \textbf{0.569} & 0.542 & 0.545 & 0.545 & 0.528 \\
    Global LV dysfunction & 0.741 & 0.713 & \textbf{0.742} & 0.735 & 0.405 \\
    \midrule
    \textbf{Macro-F1} & \textbf{0.703} & 0.688 & 0.690 & 0.693 & 0.608 \\
    \bottomrule
  \end{tabular}
\end{table}

\begin{figure}[t]
  \centering
  \resizebox{0.8\linewidth}{!}{%
  \begin{tikzpicture}[x=27cm,y=0.62cm,font=\scriptsize]
    \draw[->] (-0.04,0) -- (0.12,0);
    \node[anchor=north] at (0.04,-0.6) {$\Delta$Macro-F1};
    \draw[dashed] (0,0.3) -- (0,8.7);
    \foreach \x in {-0.04,0,0.04,0.08,0.12} {
      \draw (\x,-.10) -- (\x,.10) node[below=2pt] {\x};
    }
    \node[anchor=east] at (-0.043,8) {SFT all $-$ reports OR};
    \node[anchor=east] at (-0.043,7) {LightGBM $-$ reports OR};
    \node[anchor=east] at (-0.043,6) {SFT all $-$ LightGBM};
    \node[anchor=east] at (-0.043,5) {SFT report only $-$ reports OR};
    \node[anchor=east] at (-0.043,4) {SFT all $-$ SFT report only};
    \node[anchor=east] at (-0.043,3) {SFT all $-$ all signals OR};
    \node[anchor=east] at (-0.043,2) {SFT all $-$ GPT-5.5};
    \node[anchor=east] at (-0.043,1) {SFT report only $-$ GPT-5.5};
    \draw[line width=.6pt] (-0.030,8) -- (0.004,8);
    \fill[fusiongreen] (-0.012,8) circle (1.7pt);
    \draw[line width=.6pt] (-0.040,7) -- (0.012,7);
    \fill[fusiongreen] (-0.014,7) circle (1.7pt);
    \draw[line width=.6pt] (-0.026,6) -- (0.030,6);
    \fill[fusiongreen] (0.002,6) circle (1.7pt);
    \draw[line width=.6pt] (-0.026,5) -- (0.006,5);
    \fill[fusiongreen] (-0.010,5) circle (1.7pt);
    \draw[line width=.6pt] (-0.014,4) -- (0.010,4);
    \fill[fusiongreen] (-0.003,4) circle (1.7pt);
    \draw[line width=.6pt] (0.046,3) -- (0.105,3);
    \fill[fusiongreen] (0.075,3) circle (1.7pt);
    \draw[line width=.6pt] (0.056,2) -- (0.110,2);
    \fill[fusiongreen] (0.083,2) circle (1.7pt);
    \draw[line width=.6pt] (0.058,1) -- (0.113,1);
    \fill[fusiongreen] (0.085,1) circle (1.7pt);
    \node[anchor=west] at (0.123,8) {$-0.012$ [$-0.030$, 0.004]};
    \node[anchor=west] at (0.123,7) {$-0.014$ [$-0.040$, 0.012]};
    \node[anchor=west] at (0.123,6) {$+0.002$ [$-0.026$, 0.030]};
    \node[anchor=west] at (0.123,5) {$-0.010$ [$-0.026$, 0.006]};
    \node[anchor=west] at (0.123,4) {$-0.003$ [$-0.014$, 0.010]};
    \node[anchor=west] at (0.123,3) {$+0.075$ [0.046, 0.105]};
    \node[anchor=west] at (0.123,2) {$+0.083$ [0.056, 0.110]};
    \node[anchor=west] at (0.123,1) {$+0.085$ [0.058, 0.113]};
  \end{tikzpicture}
  }
  \caption{Exploratory paired patient-cluster bootstrap differences in Macro-F1. Points show
    paired differences, horizontal lines show 95\% CIs, and numeric annotations give the point
    estimate followed by the CI. Analyses use fixed OOF predictions.}
  \label{fig:performance}
\end{figure}
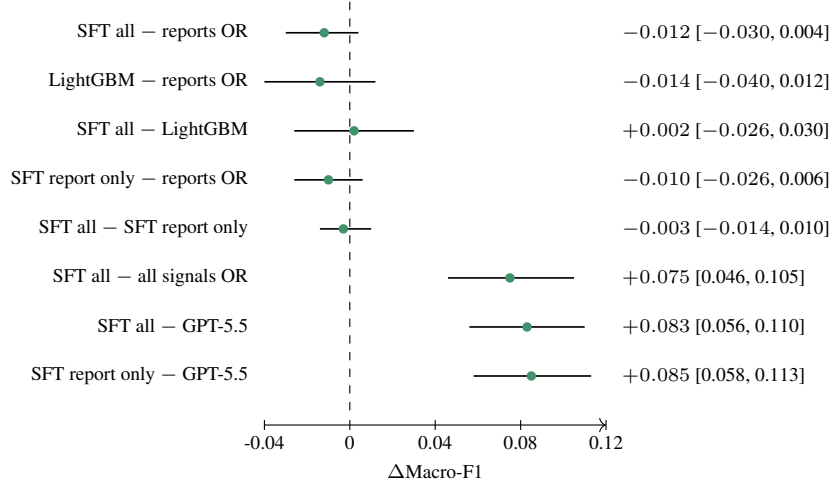

\paragraph{A supervised LLM matches conventional learned fusion and a strong fixed fusion strategy.}
The proposed LLM (SFT, all) achieved a Macro-F1 of 0.690. Among the fixed reference strategies,
reports OR performed best, with a Macro-F1 of 0.703. LightGBM, the conventional learned fusion
method, achieved a Macro-F1 of 0.688 (Table~\ref{tab:overall}).

The Macro-F1 difference between SFT all and LightGBM was $+0.002$, and the 95\% CI from the
exploratory paired patient-cluster bootstrap included zero ([$-0.026$, 0.030];
Figure~\ref{fig:performance}). The difference between SFT all and reports OR was likewise small, and
its CI included zero. These results do not establish statistical equivalence or non-inferiority, but
the point estimates indicate that the supervised LLM reached fusion performance close to both the
conventional learned fusion method and the strong fixed reference strategy evaluated here.

LightGBM required a fixed-length schema designed in advance, including mappings between targets and
view-specific report and measurement inputs, one-hot encoding of report classes, and a separate
classifier for each target. Reports OR similarly required such mappings and prior binarization of
each output. In contrast, SFT all was not given explicit features mapping individual outputs to
evaluation targets and instead processed finding statements and continuous values within a shared
text context. Its performance under this condition was comparable to LightGBM and reports OR,
indicating that language-based LLM fusion without target-specific structured feature engineering
can achieve performance similar to conventional learned fusion.

The ranking of fixed rules also depended on the metric. Reports AND had lower Macro-F1 but the
highest exact match and Hamming accuracy (Table~\ref{tab:overall}). Because negative examples
dominated this dataset, a conservative AND rule benefited metrics that reward correctness across
all targets or decisions, whereas reports OR better detected positive classes and therefore
achieved higher Macro-F1.

No single method consistently dominated across targets. Compared with LightGBM, SFT all was
slightly higher for LA dilatation, aortic root dilatation, and global LV systolic dysfunction, but
lower for LV dilatation. Compared with reports OR, SFT all was higher for LA dilatation and global
LV systolic dysfunction but lower for LV dilatation and aortic root dilatation
(Table~\ref{tab:targetf1}).

\paragraph{Learned fusion avoids degradation from indiscriminately adding weak measurements.}

Report predictions were the dominant upstream signals in this study; reference strategies that used
measurements, such as all signals OR, performed worse than report-only strategies such as reports OR
(Table~\ref{tab:overall}). Measurements could therefore behave as noise rather than complementary
information and impair SFT all. To examine this possibility, we trained \textbf{LLM (SFT, report
only)} as a diagnostic ablation. Its training conditions, system prompt, and output schema matched
SFT all, but measurement inputs were removed. Similar performance would indicate that SFT all did
not indiscriminately use weak measurements and could suppress their influence.

SFT all and SFT report only had nearly identical overall performance, and the CI for their paired
difference included zero (Table~\ref{tab:overall} and Figure~\ref{fig:performance}). At the target
level, adding measurements increased recall for LA dilatation and global LV systolic dysfunction,
but
precision--recall trade-offs kept the F1 differences small. SFT all was slightly lower for LV
dilatation, and the two models made identical predictions for every examination for aortic root
dilatation (Table~\ref{tab:targetf1}).

Moreover, indiscriminately applying OR to all signals yielded a Macro-F1 of 0.615, whereas SFT all
and LightGBM achieved 0.690 and 0.688, respectively. Thus, indiscriminately adding measurements with
low standalone performance can degrade fusion performance, while both supervised learned fusion
methods avoided this degradation.

\paragraph{Task-specific learning yields more appropriate fusion than a general-purpose LLM.}

GPT-5.5 without additional training had lower Macro-F1 and exact match than SFT all, and the CI for
the paired difference excluded zero (Table~\ref{tab:overall} and Figure~\ref{fig:performance}).

The target-specific F1 difference was especially large for global LV systolic dysfunction
(Table~\ref{tab:targetf1}). GPT-5.5 produced many false positives alongside high recall, whereas
SFT all suppressed false positives and achieved a better precision--recall balance. The
general-purpose LLM appeared to respond too strongly in an effort to avoid missing a finding, while
SFT all adapted its decisions to the upstream models' error characteristics.

This finding indicates that, even with a strong LLM, general medical knowledge and prompt-specified
decision definitions alone may be insufficient to account for upstream-model-specific reliability
and correlated errors, and that supervised learning from examination labels is useful for acquiring
an appropriate fusion strategy.

\section{Conclusion and limitations}

Supervised LLM late fusion processed multi-view report predictions and measurement predictions in
one framework without hand-designed, target-specific fusion rules. The proposed model performed
comparably to both reports OR, the best fixed rule evaluated, and LightGBM, a conventional learned
fusion method using a fixed-length feature schema, while outperforming naive fusion that
indiscriminately incorporated weak inputs and a general-purpose LLM without task-specific training.
Its performance also remained comparable to a report-only model, indicating that it avoided the
degradation caused by indiscriminately adding measurements when report predictions dominated. These
findings support the potential of language-based fusion without target-specific structured feature
engineering to achieve performance similar to the conventional learned fusion method evaluated
here.

This study has several limitations. First, it used only retrospective, single-center data;
generalization to external institutions and future time periods was not evaluated. Second, only 77
examinations were positive for LV dilatation and 70 for global LV systolic dysfunction, resulting
in wide CIs for some estimates. Third, we did not vary the number or configuration of input models,
simulate missing views, or evaluate robustness to perturbations. The claimed extensibility is
therefore currently a design-level advantage rather than an empirically demonstrated property.
Fourth, LightGBM was evaluated with a single hyperparameter configuration fixed in advance and
therefore does not represent the upper bound of performance attainable by conventional fusion.
Validation with different input configurations, additional conventional meta-learners, and external
data is needed.

\bibliographystyle{plainnat}
{\small\bibliography{references}}

\appendix

\section{Cohort characteristics}
\label{app:cohort}

Table~\ref{tab:cohort} summarizes the sex and age distributions of the 859 examinations. Because
some patients contributed more than one examination, these are examination-level summaries.

\begin{table}[h]
  \caption{Examination-level cohort characteristics.}
  \label{tab:cohort}
  \centering
  \small
  \begin{tabular}{lr}
    \toprule
    Characteristic & Summary \\
    \midrule
    Sex, male, $n$ (\%) & 496 (57.7\%) \\
    Sex, female, $n$ (\%) & 363 (42.3\%) \\
    Age, median [IQR], years & 63.0 [49.0--74.0] \\
    Age, range, years & 8--93 \\
    \bottomrule
  \end{tabular}
\end{table}

\section{Target-specific input signals}
\label{app:signals}

\begin{table}[h]
  \caption{Report and measurement inputs associated with each evaluation target.}
  \label{tab:signals}
  \centering
  \small
  \begin{tabular}{llll}
    \toprule
    Target & Report inputs & Measurement inputs & Positive direction \\
    \midrule
    LV dilatation & 4CH, 2CH, LAX & $D_d$ & Larger \\
    LA dilatation & 4CH, 2CH, LAX & LAD & Larger \\
    Aortic root dilatation & 3CH, LAX & Valsalva diameter & Larger \\
    Global LV dysfunction & 4CH, 2CH, 3CH, SAX & EF$_{\mathrm{MOD}}$, EF$_{\mathrm{Teichholz}}$ &
    Smaller \\
    \bottomrule
  \end{tabular}
\end{table}

\section{Prompts and output schema}
\label{app:prompts}

The system prompt was:
\begin{verbatim}
You are an expert echocardiography interpreter.

Determine whether each condition is positive or negative.

- Classify each condition as negative if normal, absent, trace,
  borderline, or low-normal; otherwise, classify it as positive.
- Some predictions in given reports and measurements may conflict.

Output in the following format:

<answer>
{
  "LV dilatation": "positive|negative",
  "LA dilatation": "positive|negative",
  "Aortic root dilatation": "positive|negative",
  "Global LV systolic dysfunction": "positive|negative"
}
</answer>
\end{verbatim}

An example user prompt for LLM (SFT, all) was:
\begin{verbatim}
Predicted echo report:
[4CH]
No left ventricular enlargement. Normal left ventricular wall 
motion. Mild left atrial enlargement present.
[2CH]
No left ventricular enlargement. Normal left ventricular wall 
motion. Mild left atrial enlargement present.
[3CH]
Normal left ventricular wall motion. No aortic dilatation.
[LAX]
No left ventricular enlargement. Mild left atrial enlargement 
present. No aortic dilatation.
[SAX]
Normal left ventricular wall motion.

Measurements predicted from LAX:
- EF(Teichholz)[%]: 54
- Valsalva[cm]: 3.7
- LAD[cm]: 3.8
- Dd[cm]: 4.7

Biplane Simpson (MOD) measurement predicted from 2CH and 4CH:
- EF[%]: 66
\end{verbatim}

For LLM (SFT, report only), the input ended after the five view-specific finding-statement blocks
under \texttt{Predicted echo report:}. The teacher completion and inference output used the
\texttt{<answer>} tag and four-target JSON object shown in the system prompt.

\section{Supervised fine-tuning configuration}
\label{app:hyperparameters}

\begin{table}[h]
  \caption{Supervised fine-tuning configuration.}
  \label{tab:hyperparameters}
  \centering
  \small
  \begin{tabular}{ll}
    \toprule
    Hyperparameter & Setting \\
    \midrule
    Base model & \texttt{google/medgemma-4b-it} \\
    LoRA rank / alpha / dropout & 8 / 16 / 0.05 \\
    Learning rate & $1\times10^{-5}$ \\
    Warmup ratio / scheduler & 0.05 / linear \\
    Epochs & 5 \\
    Per-GPU batch size & 4 \\
    GPUs & A100 $\times$ 4 \\
    Numerical precision & bfloat16 \\
    \bottomrule
  \end{tabular}
\end{table}

\end{document}